\documentclass[11pt]{article}
\usepackage[margin=1in]{geometry}
\usepackage{newtxtext,newtxmath}
\usepackage{microtype}
\usepackage{booktabs}
\usepackage{graphicx}
\usepackage{amsmath}
\usepackage{multirow}
\usepackage[round,authoryear]{natbib}
\usepackage{xcolor}
\usepackage[colorlinks=true,linkcolor=blue!60!black,citecolor=blue!60!black,urlcolor=blue!60!black]{hyperref}
\usepackage[font=small,labelfont=bf]{caption}

\newcommand{\pairacc}{pair acc.}
\newcommand{\partialrho}{partial $\rho$}

\title{\textbf{Verifiable but Not Predictable:\\What Sentence Embeddings Know\\About Propositional Decomposition}}
\author{
  Qingan Li \qquad Tianxiao Wu \qquad Xiling Cui\\[0.3em]
  Technological and Higher Education Institute of Hong Kong, Hong Kong SAR\\[0.3em]
  \small\texttt{qingan.li@thei.edu.hk}, \texttt{ng.tx.king@gmail.com}, \texttt{cuixiling@gmail.com}
}
\date{November 2026}

\begin{document}
\maketitle

\begin{abstract}
Quality control for proposition decomposition by large language models (LLMs) raises two
asymmetric questions. After decomposition, one can \emph{verify}: given a sentence and a candidate
proposition, does the proposition belong to that sentence? Before decomposition, one would like to
\emph{predict}: how many propositions should the sentence yield? On PropSegmEnt, a corpus of
English Wikipedia sentences each segmented by three human annotators, we find that the two
questions live in different representational spaces. A natural language inference (NLI) judge
verifies membership with 92.8\% accuracy and an area under the ROC curve (AUC) of 0.977, and its
residual errors turn out to be mostly propositions that are true in document context rather than
genuine entailment mistakes. Vector geometry cannot verify: cosine similarity saturates, and a
sentence embedding is not a linear combination of its propositions' embeddings. For prediction,
every symbolic anchor we test, namely token count, the CPIDR idea-density counter, and a
Neo-Davidsonian predicate count, reduces to a length proxy once token count is controlled. A
linear ridge probe on sentence embeddings is the only predictor that keeps genuine count signal:
its length-partialled correlation with human counts is 0.40--0.49 with bootstrap confidence
intervals (CIs) excluding zero, and it orders length-matched sentence pairs correctly 67--71\% of
the time against a 50\% chance level, statistically matching the inter-annotator agreement ceiling.
This signal survives INT8 quantization without loss, INT4 at a small cost, and Matryoshka
truncation to 256 dimensions, and it works as well in a 2021 encoder as in a 2025 one. Its
structured weakness is a systematic under-splitting of long sentences. We distill the protocol
lessons, including a retracted pilot study, into a checklist for probing length-confounded
quantities.
\end{abstract}

\section{Introduction}
\label{sec:intro}

Propositions are minimal, self-contained units of meaning. Decomposing sentences into propositions
has become a standard building block of modern language pipelines: proposition-level retrieval
units improve dense retrieval and cross-lingual search \citep{chen2024densex}, and
decompose-then-verify designs underpin complex claim checking and factual precision measurement
\citep{chen2022claimdecomp,min2023factscore}. Proposition segmentation itself is now scalable
enough to serve as an off-the-shelf tool \citep{hosseini2024aps}. As large language models (LLMs)
are routinely asked to perform this decomposition, a quality-control question follows: who checks
the decomposer?

The question is not hypothetical. In a benchmark of five LLM decomposers on 48 ESG disclosure
sentences that we ran as part of a prior audit, granularity varied wildly across models. The ratio
of produced propositions to a symbolic expectation had median $0.6$, one model over-split at a
ratio of $1.13$, and the best model reached only $0.599$ on a composite proposition-quality index.
No external reference existed against which to judge any individual split.

\begin{figure}[t]
\centering
\includegraphics[width=0.98\textwidth]{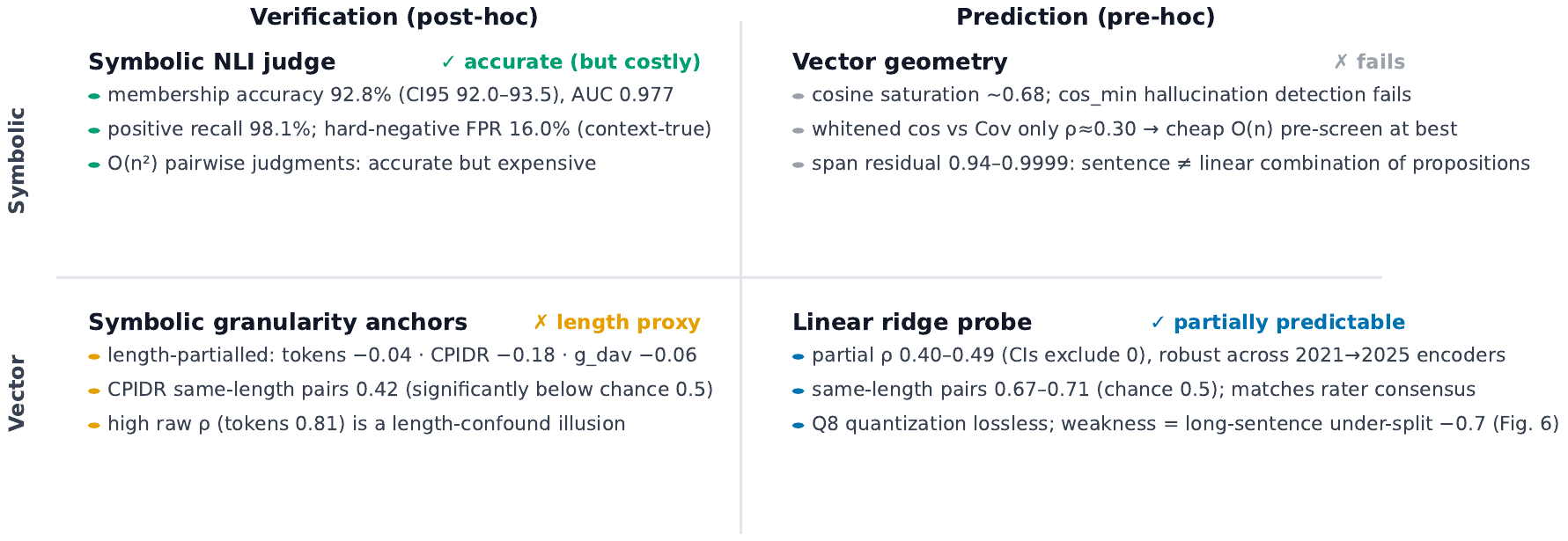}
\caption{The two directions of the sentence--proposition relation crossed with two method
families. Verification is solved symbolically but not geometrically; prediction is partially
solved by a linear embedding probe, while symbolic anchors act as length proxies. All numbers are
developed in Sections~\ref{sec:verify}--\ref{sec:analysis}.}
\label{fig:concept}
\end{figure}

This paper argues that decomposition quality control in fact contains two questions with sharply
different difficulty profiles, and that the difference between them is itself informative about
where the relevant information lives (Figure~\ref{fig:concept}). \emph{Verification} is post-hoc
and has a reference: given a sentence $s$ and a candidate proposition $p$, decide whether $p$
belongs to the decomposition of $s$. \emph{Prediction} is pre-hoc and has none: given only $s$,
estimate how many propositions a faithful decomposition should yield, that is, the expected
granularity before any expensive decomposition is run. Using PropSegmEnt
\citep{chen2023propsegment}, 6{,}093 usable English Wikipedia sentences each segmented by three
annotators, we measure both directions under one protocol and find that each direction is solved,
partially solved, or unsolved depending on the method family.

Read summarily, a symbolic natural language inference (NLI) judge verifies membership accurately,
purely geometric alternatives fail, no symbolic predictor survives length control, and a linear
embedding probe reaches the human agreement ceiling while failing in a structured way on long
sentences. Figure~\ref{fig:pipeline} in Section~\ref{sec:formulation} lays out the full study
design, and Sections~\ref{sec:verify}--\ref{sec:analysis} develop each of these four findings in
turn.

Why should others care about the asymmetry? Because it locates the two kinds of information.
Faithfulness, the object of verification, lives in symbolic inferential space and is readable by a
discriminative judge. Granularity, the object of prediction, is partially encoded in embedding
space already, and broadly so: our best probe runs on a 22M-parameter encoder from 2021. Neither
half is fully solved, and knowing which half is which prevents both over-claiming, as when a
cosine threshold is proposed as a faithfulness filter, and under-claiming, as when expected
granularity is treated as unknowable before decomposition.

\paragraph{Contributions.}
(i) A formulation of decomposition quality control as an asymmetric verify/predict pair, with a
$2\times2$ empirical map of which method family solves which direction. (ii) The first
measurements of NLI membership accuracy for propositionalization on PropSegmEnt, with an error
analysis showing that residual false positives concentrate in context-true, coreference-entailed
propositions. (iii) A length-controlled study of proposition-count decodability in sentence
embeddings, with bootstrap CIs and paired ceiling comparisons showing that the probe matches
inter-annotator agreement while every symbolic anchor reduces to a length proxy. (iv) A robustness
sweep over encoders, quantization levels, and Matryoshka dimensions. (v) A documented retraction
trajectory of our own pilot study that doubles as a protocol checklist for probing
length-correlated linguistic quantities.

\section{Related Work}
\label{sec:related}

\paragraph{Proposition segmentation and decomposition.}
The segmentation task has solid supervised footing since PropSegmEnt \citep{chen2023propsegment},
which collected human proposition-level segmentations for more than 13{,}000 sentences, over
45{,}000 propositions in total with three annotators per sentence, together with
proposition-level entailment labels against the source documents. On the applied side, Dense X
retrieval \citep{chen2024densex} showed that indexing \emph{self-contained} propositions instead
of passages improves retrieval-augmented question answering, which is what motivated LLM-based
decomposition at scale. ClaimDecomp \citep{chen2022claimdecomp} decomposes complex political
claims into yes/no subquestions whose answers determine claim veracity, and trains models to
generate such decompositions from 1{,}200 annotated examples. \citet{hosseini2024aps} train an
abstractive proposition segmentation model on LLM outputs filtered by an entailment checker,
making segmentation scalable and domain-general. Decomposition quality itself, however, is
rarely audited: \citet{wanner2024closer} show that FActScore-style evaluation is sensitive to
which decomposition method produced the claims, so metric error and decomposition error blur
together. \citet{srikanth2025atomic} decompose NLI hypotheses into atomic sub-problems and find
that LLMs still struggle with logical consistency on the atomic level. All of these works take
the \emph{decomposition} as the deliverable and evaluate it downstream. None asks the two
quality-control questions we study: whether a produced proposition can be checked against its
sentence, and whether the expected count can be known before decomposition.

\paragraph{Verification by entailment.}
Using NLI \citep{williams2018mnli,conneau2018xnli} as a non-generative faithfulness judge has
strong precedent. For summarization consistency, FactCC \citep{kryscinski2020factcc} fine-tunes a
RoBERTa encoder on automatically generated fact manipulations, and SummaC \citep{laban2022summac}
aggregates sentence-level NLI decisions into document-level consistency scores; on a benchmark
assembled from six document-level NLI datasets, its lightweight discriminative aggregations
outperform end-to-end alternatives. FActScore \citep{min2023factscore} decomposes a long-form
generation into atomic facts and verifies each against retrieval over a reference corpus,
reporting the fraction supported; applied to biographies, it found that ChatGPT supported only
58\% of its atomic facts. Our judge follows this discriminative
tradition. What is missing from these pipelines is the granularity side: they check \emph{whether}
produced units are supported, not \emph{how many} units should have been produced, and the
prediction direction of this paper targets exactly that gap.

\paragraph{Probing representations.}
Linear probes are the standard instrument for asking what a representation encodes
\citep{alain2017probes,hewitt2019structural}. For sentence embeddings specifically,
\citet{conneau2018cram} introduced ten probing tasks covering properties from sentence length to
topological constituents, and \citet{adi2016length} probed sentence length, word content, and
word order directly; that length itself was an early probe target foreshadows the confound this
paper treats as central. Probing BERT layer by layer recovers the classical NLP pipeline in order
\citep{tenney2019bert}, and selectivity critiques
\citep{hewitt2019control,belinkov2022probing,ravichander2021amnesia} warn that probes can exploit
surface statistics rather than linguistic structure. Proposition count is a nearly worst case for
this failure mode, because it correlates with token length at $\rho=0.749$ on our data. Our
response is to build the confound into the evaluation itself: we include a token-count baseline
that \emph{wins} the raw correlation, we report length-partialled correlations, and we require
decisions to hold on exact-length pairs, in the spirit of control tasks but adapted to
regression.

\paragraph{Symbolic granularity anchors.}
Proposition counting has a four-decade psycholinguistic pedigree. The Kintsch text-base model
treats a sentence's meaning as a network of propositions \citep{kintsch1978model}, and the
Turner--Greene manual \citep{turner1977construction} operationalized proposition-level coding;
CPIDR \citep{brown2008automatic} automated idea-density estimation from part-of-speech rules.
Idea density measured in early-life autobiographies predicted late-life Alzheimer pathology in
the Nun Study \citep{snowdon1996linguistic}, the strongest evidence that proposition counts carry
cognitive substance: among the sisters who died and were autopsied, all of those with low idea
density in early life had confirmed Alzheimer pathology and none of those with high density did.
The association was replicated with the automated CPIDR tool itself on medical-school essays in
the Precursors Study \citep{engelman2010precursors}. Neo-Davidsonian event semantics motivates
counting nuclear predicates \citep{davidson1967logical}. We evaluate this symbolic family
directly against the human reference
counts and find that, on modern text, it transmits no length-independent signal.

\paragraph{Embedding geometry, compression, and uncertainty.}
Post-hoc geometric corrections such as whitening \citep{su2021whitening} and flow calibration
\citep{li2020bertflow} improve embedding isotropy for similarity tasks, and contrastive training
\citep{gao2021simcse} is the dominant modern recipe. Granularity itself has begun to attract
attention from the embedding side: the sub-sentence encoder \citep{chen2023subsentence}
contrastively learns proposition-level embeddings rather than sentence-level ones,
\citet{xu2025granularity} document a granularity dilemma in which encoders fail to match
fine-grained entities and events inside a sentence, and Granuscore
\citep{ellinger2026granuscore} measures text granularity reference-free from hierarchical
embedding structure; none of these predicts the human proposition count of a given sentence,
which is our target. Separately, Matryoshka representation
learning \citep{kusupati2022matryoshka} packs nested granularities into a single embedding so
that leading dimensions alone remain accurate, and \citet{dettmers2023int4} showed that 4-bit
quantization matches 16-bit accuracy for frozen LLMs across dozens of zero-shot tasks; both
results frame our robustness sweeps, and both behave as predicted under count probing. Closest in
spirit to our question, semantic entropy \citep{farquhar2024detecting} detects LLM hallucinations
by clustering sampled answers into bidirectional-entailment equivalence classes and measuring
entropy over cluster sizes, outperforming token-level uncertainty across multiple open-domain QA
benchmarks, and semantic-entropy probes \citep{kossen2024semantic} train linear probes on hidden
states to predict that entropy without sampling at inference time. These methods probe model
\emph{uncertainty}; our target is a structural property of the \emph{input}, namely how many
meaning units it contains, and it turns out to be broadly decodable.

\section{Problem Formulation: Verify vs.\ Predict}
\label{sec:formulation}

\begin{figure}[t]
\centering
\includegraphics[width=0.98\textwidth]{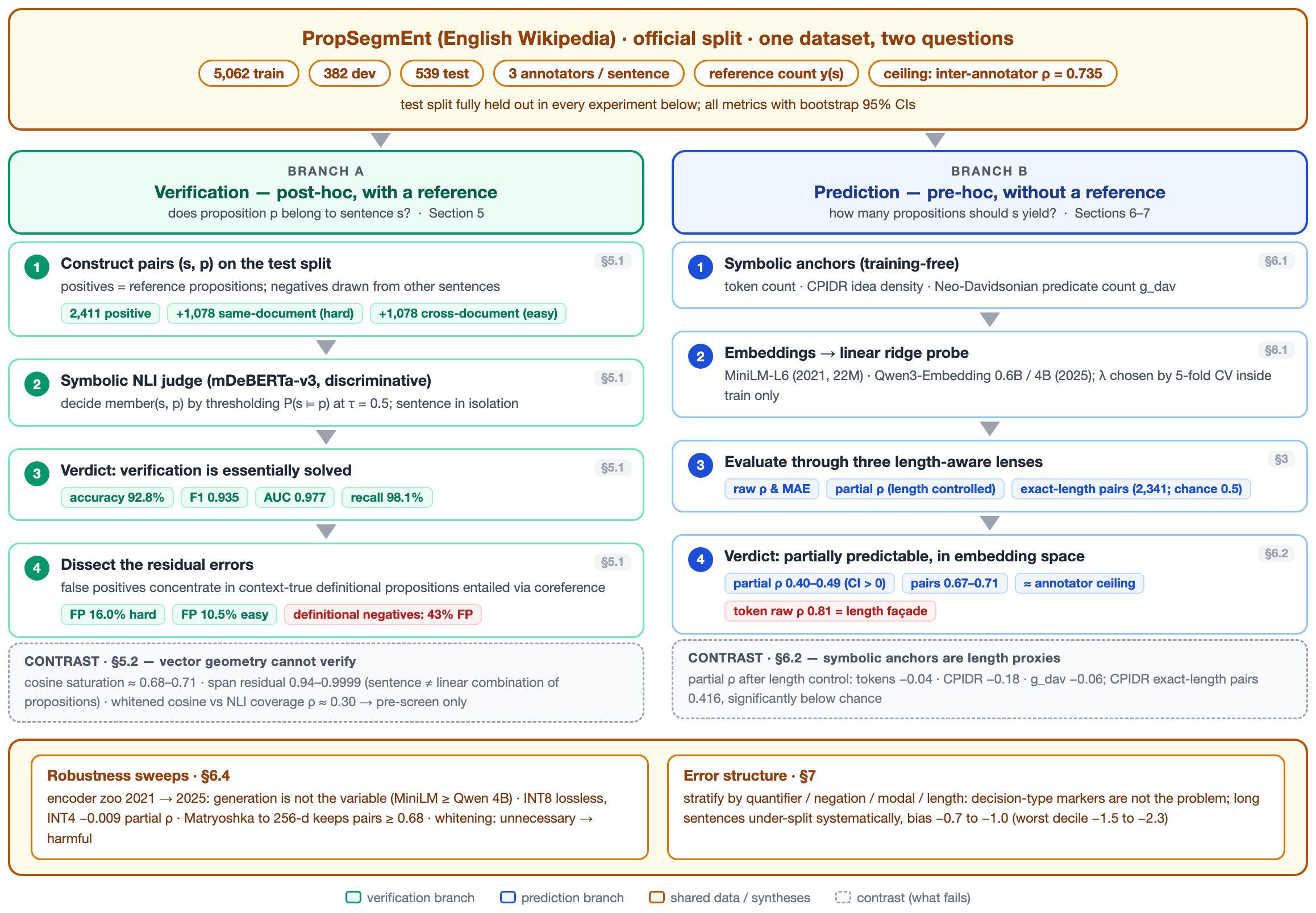}
\caption{Study design. One dataset feeds two branches. Branch A (verification) constructs
$(s,p)$ pairs and applies an NLI judge, then dissects its errors. Branch B (prediction) pits
symbolic anchors against linear ridge probes on sentence embeddings, evaluated through three
complementary lenses against the inter-annotator ceiling. Both branches feed a shared robustness
and error-structure layer. Section numbers mark where each stage is reported.}
\label{fig:pipeline}
\end{figure}

The introduction posed two questions of unequal difficulty. Before measuring either, this section
fixes the objects, the two tasks, the metrics, and the references against which ``good'' is
judged. Figure~\ref{fig:pipeline} summarizes the resulting design: one dataset feeds a
verification branch and a prediction branch that share split hygiene, and the two meet again in a
shared robustness and error-structure layer.

The design is asymmetric by intent, because the two tasks fail in different ways and therefore
need different controls. Verification is a classification problem whose risk is a misleadingly
high accuracy on easy negatives, so we build hard negatives and dissect false positives.
Prediction is a regression problem whose risk is a length proxy masquerading as a counter, so we
build length into the evaluation itself through the three lenses defined below. Each lens is
cheap to game alone, but together they are hard to fool, and we require every conclusion to hold
across all three.

\subsection{Objects}
Let $s$ be a sentence and $D(s)=\{p_1,\dots,p_k\}$ a decomposition into propositions. PropSegmEnt
realizes $D(s)$ as a segmentation of $s$ into token spans, which may be non-contiguous; each
sentence carries three annotator segmentations, one of which is flagged as the reference. Write
$y(s)$ for the \emph{reference count} and $y_1,y_2,y_3$ for the three annotator counts.

\subsection{Task V: verification}
\begin{figure}[t]
\centering
\includegraphics[width=0.98\textwidth]{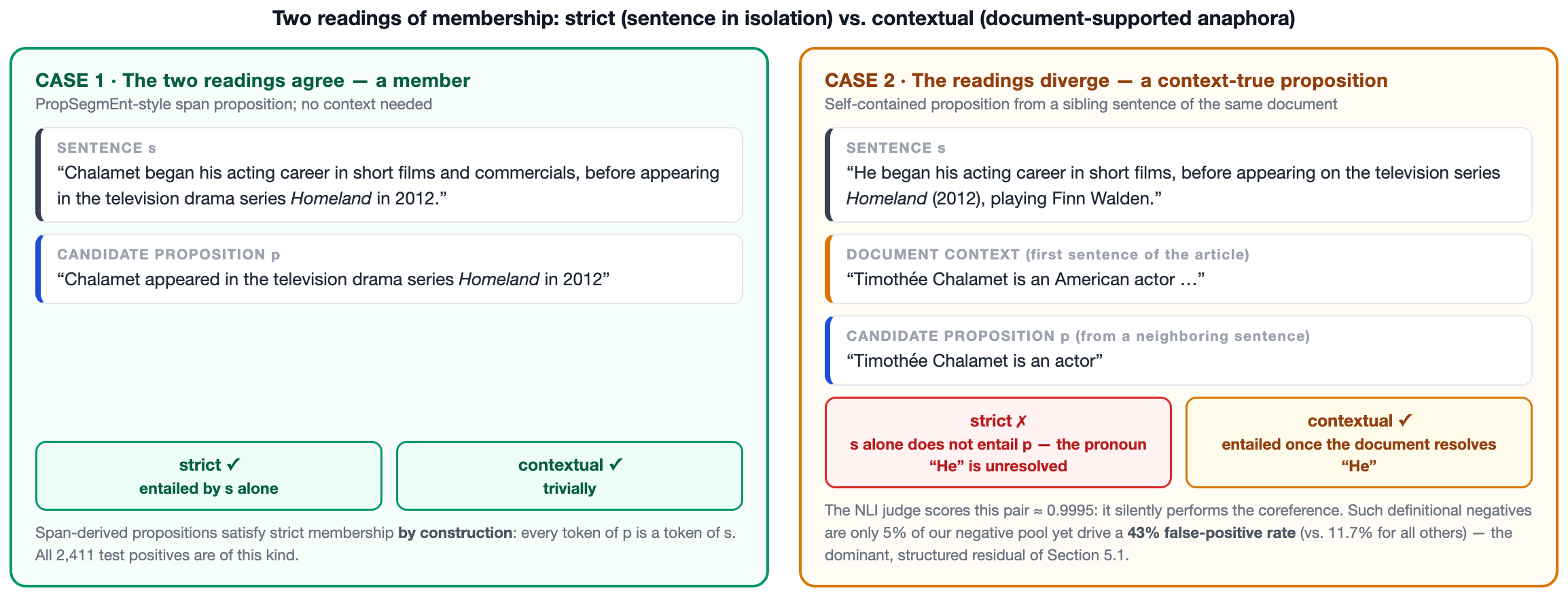}
\caption{Two readings of membership, illustrated with test-set pairs. \emph{Left}: the
proposition restates a span of the sentence with the reference made explicit, and both readings
accept it. \emph{Right}: the sentence alone cannot entail the definitional proposition, because
its subject appears only as an unresolved pronoun; strict membership therefore rejects what
contextual membership, given the document, accepts. The NLI judge scores this pair $0.9995$, and
definitional negatives of this shape are 5\% of the negative pool yet drive 43\% of the false
positives (Section~\ref{sec:nli}).}
\label{fig:readings}
\end{figure}

Given a pair $(s,p)$, decide \textsc{member}$(s,p)$, that is, whether $p$ faithfully represents
part of $s$'s content. Membership deserves one careful word, because it is \emph{not} identical to
strict entailment. A decomposer typically produces self-contained propositions, a
decontextualization step in the sense of \citet{choi2021decontextualization}, which may resolve
an anaphor or an ellipsis using information supplied by the surrounding document; the sentence
alone then fails to entail $p$ even though $p$ correctly expresses part of $s$'s meaning.
Conversely, a proposition lifted from a neighboring sentence may be entailed by $s$ once document
context is taken into account, without being part of $s$'s content at all. We therefore
distinguish \emph{strict membership}, the relation ``$s$ entails $p$ in isolation'' that our judge
implements, from \emph{contextual membership}, which allows document-supported anaphora
resolution. Our test positives are span-derived and satisfy strict membership by construction,
and the negative construction together with the false-positive analysis in Section~\ref{sec:nli}
measures how far the two readings drift apart in practice. Operationally, the judge scores
$P_{\mathrm{NLI}}(s \vDash p)$, and membership is declared at a threshold $\tau=0.5$.
Figure~\ref{fig:readings} makes the distinction concrete with two test-set pairs, including the
coreference pattern on which the two readings diverge.

\subsection{Task P: prediction}
Given $s$ alone, produce an estimate $\hat y(s)\in\mathbb{R}$ of the reference count and evaluate
against $y(s)$ under three complementary lenses:
\begin{enumerate}
\item \emph{Raw} Spearman $\rho$ and mean absolute error (MAE) over test sentences;
\item \emph{Length-partialled} $\rho$: the rank-residual correlation between $\hat y$ and $y$
after regressing both on token count, which answers whether there is count information beyond
length;
\item \emph{Exact-length pair accuracy}: over all test pairs of sentences with \emph{identical}
token count but different reference counts (2{,}341 pairs), the fraction where $\hat y$ orders
them correctly. Chance is 0.5, and the token-count baseline is \emph{undefined} here because it
never breaks ties, which is precisely the point.
\end{enumerate}

\subsection{Ceiling}
The construct ``number of propositions'' is human-defined, so its noise ceiling on the test split
is the agreement between annotators, measured over all 1{,}617 annotator pairs: $\rho=0.735$ with
a 95\% confidence interval (CI) of $[0.696, 0.770]$, and a pooled annotator-pair MAE of 2.03 with
CI $[1.83, 2.24]$. Any predictor's agreement with the vetted reference annotation should be read
against this inter-annotator ceiling, not against a perfect 1.0.

\section{Data, Anchors, Encoders, and Judge}
\label{sec:data}

\paragraph{PropSegmEnt.}
We use the Wikipedia portion of the dataset, because the news portion's source texts are not
redistributed: 6{,}093 usable sentences after dropping 110 flagged as non-informational or not
understood, split officially into 5{,}062 training, 382 development, and 539 test sentences, with
no document-cluster overlap. Reference counts average 4.18 with median 3. Two dataset
characteristics matter for interpretation. First, annotators mark spans of the original sentence,
and 67.9\% of all propositions are non-contiguous, reflecting a compressed annotation style.
Second, the construct is ``distinct unit of meaning,'' a close cousin but not an exact match of
the self-contained propositions of Dense X \citep{chen2024densex}; both points return in
Section~\ref{sec:limitations}. Sentence texts are CC BY-SA 3.0, and we pin dataset commit
\texttt{d5172f9} for reproducibility.

\paragraph{Symbolic anchors.}
Three training-free, deterministic counters span the symbolic families. The token count serves as
the \emph{token-count baseline}. CPIDR \citep{brown2008automatic}, the Computerized Propositional
Idea Density Rater, estimates idea density through part-of-speech rules; we use a faithful English
reimplementation over NLTK perceptron tags. The anchor g\_dav counts nuclear predicates in the
Neo-Davidsonian sense \citep{davidson1967logical}: finite verbs plus event nominalizations, with
clause-level rescue rules.

\paragraph{Encoders and judge.}
The three encoders were chosen to span era, scale, and training objective rather than to
cherry-pick a winner. MiniLM-L6-v2 \citep{wang2020minilm,reimers2019sentence} is a 22M-parameter
mean-pooled distilled encoder from 2021 that represents the lightweight pre-contrastive era.
Qwen3-Embedding 0.6B and 4B \citep{qwen3embedding} are current open-weight contrastive encoders
with published Matryoshka support and locally runnable fp16 weights, both of which our robustness
sweeps require. The membership judge is mDeBERTa-v3-base fine-tuned on multilingual NLI
\citep{he2023debertav3,fine-tuned-model-card,conneau2018xnli}. A discriminative classifier makes
verification non-generative and immune to prompt-format confounds, its multilingual training
matches our corpus aspirations, and its MIT license and modest size make it a practical default.
We use fp16 encoder weights throughout so that the reference condition carries no quantization
artifacts, and we measure quantization explicitly in Section~\ref{sec:robust}.

\section{The Verification Direction: Membership Is (Symbolically) Easy}
\label{sec:verify}

We first measure the direction that has a reference to check against.
Section~\ref{sec:nli} shows that an NLI judge verifies membership accurately, and
Section~\ref{sec:geometry} shows that purely geometric alternatives cannot, which together fix
the first row of Figure~\ref{fig:concept}.

\subsection{An NLI judge verifies membership at 92.8\% accuracy}
\label{sec:nli}

We construct 4{,}567 pairs $(s,p)$ on the test split. Positives are all 2{,}411 propositions of
the reference segmentations, reconstructed faithfully from token indices with skipped ranges
joined by spaces. Hard negatives are 1{,}078 reference propositions of \emph{other sentences in
the same document}: they are topically adjacent, which makes them hard, but they are not part of
the target sentence's content. Easy negatives are 1{,}078 propositions from other document
clusters. The judge scores $P_{\mathrm{NLI}}(s\vDash p)$, and membership is declared at
$\tau=0.5$.

\begin{table}[t]
\centering\small
\begin{tabular}{lcc}
\toprule
 & Value & 95\% CI \\
\midrule
Accuracy & \textbf{92.8\%} & [92.0, 93.5] \\
F1 (positive = member) & 0.935 & --- \\
AUC (all negatives) & \textbf{0.977} & --- \\
AUC vs.\ hard negatives & 0.970 & --- \\
Positive recall & 98.1\% & --- \\
False-positive rate, hard negatives & 16.0\% & [13.9, 18.2] \\
False-positive rate, easy negatives & 10.5\% & [8.7, 12.4] \\
Mean score, positives / hard / easy & 0.958 / 0.201 / 0.162 & --- \\
\bottomrule
\end{tabular}
\caption{Membership verification with the mDeBERTa-v3 NLI judge ($n=2{,}411$ positives,
$1{,}078$ hard + $1{,}078$ easy negatives; $\tau=0.5$; 2{,}000-resample bootstrap CIs). The
positive recall independently reproduces the soundness score of 0.982 that our pipeline computes
on the same gold data, a useful cross-check.}
\label{tab:membership}
\end{table}

\begin{figure}[t]
\centering
\includegraphics[width=0.98\textwidth]{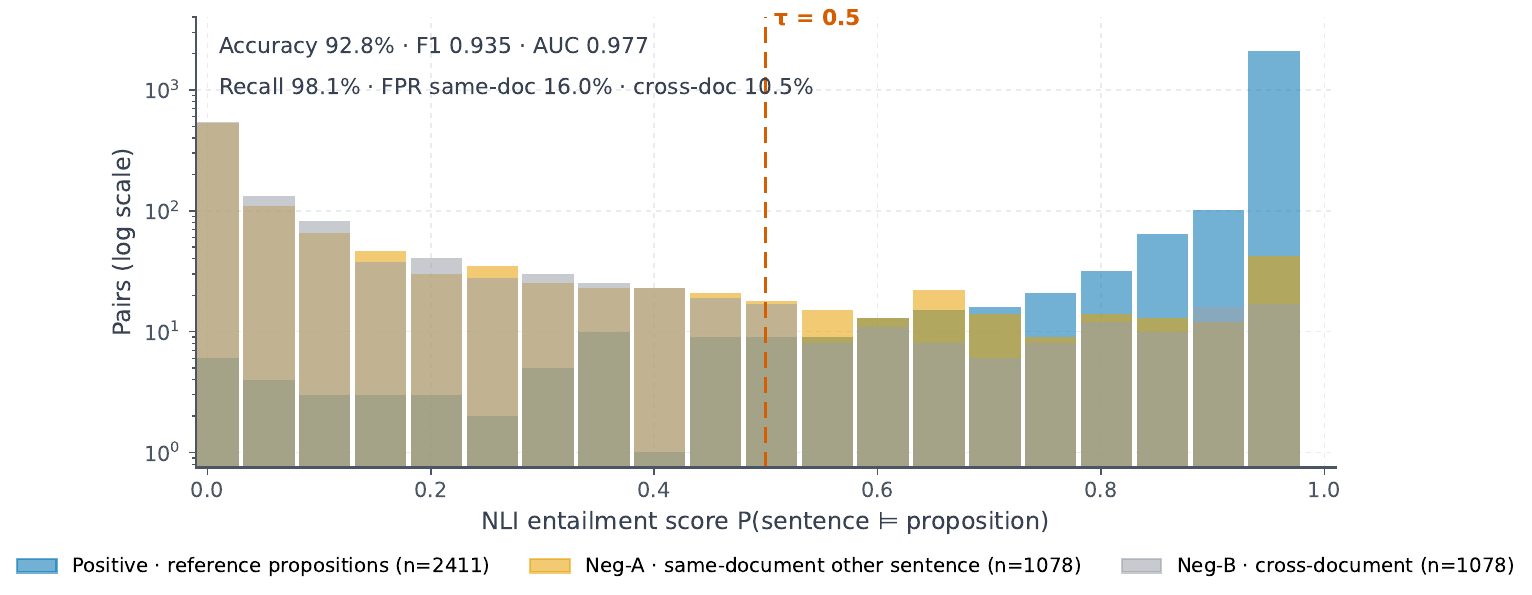}
\caption{Membership-score distributions of the NLI judge. Positives concentrate above 0.9 and
negatives below 0.2. The residual overlap is dominated by context-true definitional propositions
from sibling sentences (Section~\ref{sec:nli}).}
\label{fig:membership}
\end{figure}

Table~\ref{tab:membership} and Figure~\ref{fig:membership} give the results. Verification is, for
practical purposes, solved symbolically: 92.8\% accuracy, near-perfect ranking with AUC 0.977,
and clean score separation of 0.958 for positives against 0.16--0.20 for negatives. Two
qualifications keep the claim honest.

\paragraph{The errors that remain are structured.}
Among the 285 false positives, negatives matching the definitional pattern ``$X$ is a/an
\ldots'' make up only 5\% of negatives, drawn typically from the first sentence of a sibling
Wikipedia article, yet they produce a 43\% false-positive rate versus 11.7\% for all other
negatives. Inspection shows that they are context-true: the target sentence refers to the entity
by name or pronoun, as in \emph{``He began his acting career in short films \ldots''}, and the
judge, correctly given world knowledge and coreference, entails \emph{``Timoth\'ee Chalamet is an
actor''} (Figure~\ref{fig:readings}, right). Under strict membership these are errors; under contextual membership they are not. The
drift between the two readings, anticipated in Section~\ref{sec:formulation}, is therefore real
but concentrated: it lives mainly in definitional propositions whose subject is recoverable from
document context.

\paragraph{Accuracy is affordable only in batch.}
The judge is fast per pair, but verifying a full decomposition costs one judgment per
proposition, and screening a corpus means judging every candidate pair. The cheap vector
alternative evaluated next is precisely the pre-screen that cannot replace the judge.

\subsection{Vector geometry does not verify}
\label{sec:geometry}

If membership were linearly accessible from embeddings, a cheap vector screen could replace the
NLI judge. We piloted five geometric constructions on 48 stratified sentences with five
decomposition models and re-checked the survivors; they fail in instructive ways.

The strongest test is span reconstruction: project the sentence embedding onto the subspace
spanned by its propositions' embeddings. The residual is 0.94 under MiniLM and 0.9999 under
Qwen3, so the sentence embedding is not a linear combination of its proposition embeddings.
Direct similarity is no better. Raw cosine saturates at 0.68--0.71 for members and non-members
alike, which kills hallucination detection by minimum cosine; after corpus-level whitening,
cosine tracks the NLI coverage score at only $\rho\approx0.30$ ($n=212$, $p<0.001$), enough for a
pre-screen, not a verdict. Two further constructions fail by turning into length meters. The
spectral dimension of the proposition embedding family correlates with symbolic counts at
moderate strength, but once token count is controlled the correlation collapses to 0.07--0.09.
Greedy coverage curves, which measure how many propositions are needed to cover a given fraction
of the sentence direction, correlate with counts at 0.72--0.80, yet never exceed the raw count
$n_{\text{props}}$ itself at 0.822, so they add no count information; true final coverage is only
0.08--0.14.

Together with Section~\ref{sec:nli}, this fills the first row of Figure~\ref{fig:concept}: the
information needed for verification lives in symbolic inferential space, is accurately read by an
NLI judge, and is not linearly recoverable from current sentence embeddings.

\section{The Prediction Direction: Count Is Only Partially Predictable}
\label{sec:predict}

We now turn to the direction without a reference. Section~\ref{sec:probe} fixes the probe
protocol, Section~\ref{sec:main} reports the main comparison against symbolic anchors and the
inter-annotator ceiling, Section~\ref{sec:retraction} documents a retracted pilot whose lessons
shaped the protocol, and Section~\ref{sec:robust} sweeps encoders, quantization levels, and
embedding dimensions.

\subsection{Probe protocol}
\label{sec:probe}
For each encoder we embed all sentences, cache embeddings by the SHA-256 hash of the text, and
unit-normalize them. We standardize features and fit a ridge regression \citep{hoerl1970ridge}
from the 5{,}062 training embeddings to reference counts. The ridge penalty $\lambda$ is selected
by 5-fold cross-validation \emph{inside the training set} only; the development and test splits
are untouched. We report both the raw embedding space and a whitened space computed with
statistics fitted on train only. All metrics are computed on the 539 test sentences, with 95\%
CIs from sentence-level bootstraps, or pair-level bootstraps for pair accuracy, using
1{,}000--2{,}000 resamples. The protocol is deterministic: re-running it reproduces the headline
numbers bit-exactly, and test $\rho$ varies by less than 0.01 across two orders of magnitude of
$\lambda$ around the selected value, so no conclusion rests on penalty selection.

\begin{table}[t]
\centering\small
\begin{tabular}{llcccc}
\toprule
 & Predictor & $\rho$ & MAE & \partialrho & \pairacc \\
\midrule
\multirow{3}{*}{\rotatebox{90}{probe}}
 & MiniLM (2021, 22M) & \textbf{0.827} {[}0.795, 0.854{]} & \textbf{1.36} & \textbf{0.490} {[}0.414, 0.556{]} & 0.698 {[}0.679, 0.717{]} \\
 & Qwen3-Emb 0.6B (2025) & 0.768 {[}0.723, 0.803{]} & 1.53 & 0.399 {[}0.317, 0.472{]} & 0.669 {[}0.651, 0.688{]} \\
 & Qwen3-Emb 4B (2025) & 0.789 {[}0.748, 0.824{]} & 1.48 & 0.455 {[}0.372, 0.527{]} & \textbf{0.710} {[}0.691, 0.728{]} \\
\midrule
\multirow{3}{*}{\rotatebox{90}{symbolic}}
 & Token count & 0.810 {[}0.776, 0.840{]} & --- & $-$0.042 {[}$-$0.070, 0.068{]} & undefined \\
 & CPIDR & 0.659 {[}0.606, 0.706{]} & 6.21 & $-$0.179 {[}$-$0.262, $-$0.086{]} & 0.416 {[}0.395, 0.436{]} \\
 & g\_dav & 0.570 {[}0.509, 0.628{]} & 2.16 & $-$0.064 {[}$-$0.152, 0.029{]} & 0.483 {[}0.459, 0.507{]} \\
\midrule
 & \emph{Inter-annotator ceiling} & \emph{0.735} {[}0.696, 0.770{]} & \emph{2.03} & --- & --- \\
\bottomrule
\end{tabular}
\caption{Main results on the PropSegmEnt test split ($n=539$; 2{,}341 exact-length pairs).
Probes are ridge regressions on raw sentence embeddings; whitened variants appear in
Table~\ref{tab:robust}. The token count's MAE is in token units and omitted as incomparable, and
its pair accuracy is undefined because it never breaks length ties. CIs are bootstrap 95\%. The
ceiling is inter-annotator agreement over 1{,}617 annotator pairs.}
\label{tab:main}
\end{table}

\subsection{Main results}
\label{sec:main}
\begin{figure}[t]
\centering
\includegraphics[width=0.98\textwidth]{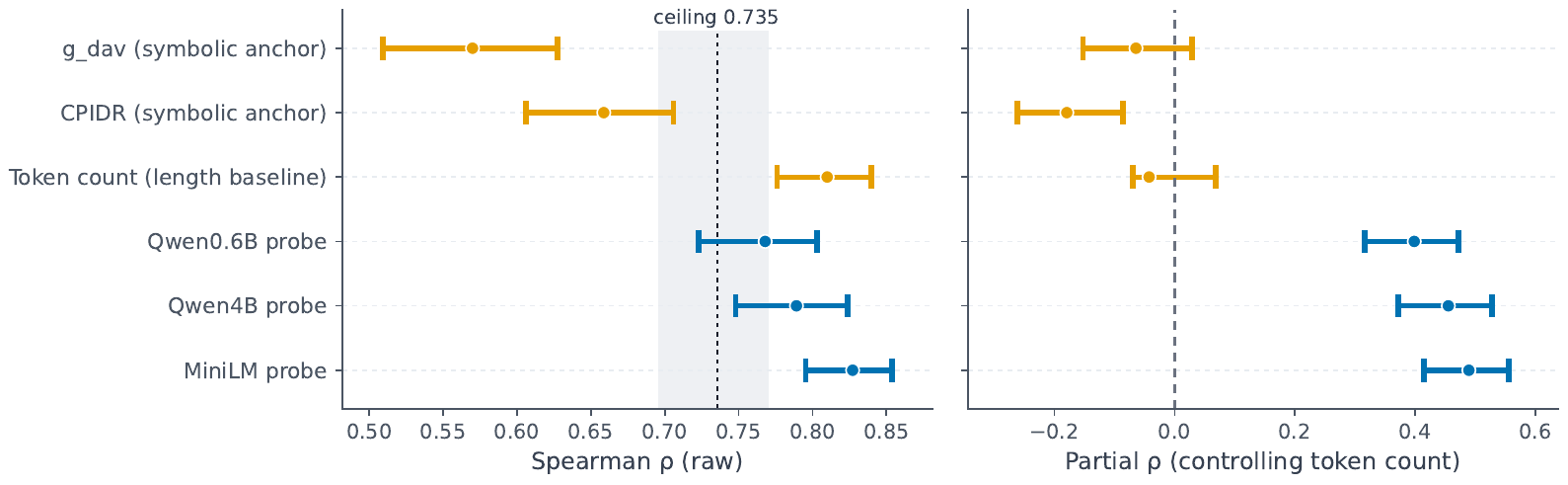}
\caption{Forest plots with bootstrap 95\% CIs. Left: raw Spearman $\rho$ against the reference
annotation, with the grey band marking the inter-annotator ceiling CI and black dashes its point
estimate. Right: partial $\rho$ controlling token count. Symbolic anchors collapse to or below
zero under length control; probes hold 0.40--0.49.}
\label{fig:forest}
\end{figure}

Table~\ref{tab:main} and Figure~\ref{fig:forest} support three statements.

\paragraph{Only the probe transmits count information beyond length.}
All three probes have partial $\rho$ CIs entirely above zero, the weakest at 0.317. The token
count's partial CI straddles zero at $[-0.070, 0.068]$, CPIDR's is significantly negative, and
g\_dav's straddles zero. On exact-length pairs, where length cannot help, CPIDR discriminates at
0.416 with CI $[0.395, 0.436]$, significantly below the 0.5 chance level, while the probes hold
0.669--0.710. The symbolic-anchor family, despite its psycholinguistic pedigree, acts as a set of
length proxies on this benchmark.

\paragraph{Raw correlation would have lied.}
The token count attains raw $\rho=0.810$, above two of the three probes, and the paired
difference against the 0.6B probe is genuinely negative at $-0.042$ with CI $[-0.081, -0.006]$.
MiniLM is the only probe that also wins the raw comparison at 0.827. This is the length confound
in its purest form, and it is the reason the protocol mandates the three lenses jointly
(Section~\ref{sec:formulation}).

\paragraph{The probe matches, but does not exceed, human consensus.}
Against the vetted reference annotation, MiniLM's $\rho=0.827$ nominally exceeds the 0.735
inter-annotator ceiling. The comparison is subtle, because the ceiling averages noisy annotator
pairs while the probe predicts a cleaned reference. The paired bootstrap difference between probe
and ceiling is $+0.032$ with CI $[-0.012, 0.073]$, statistically indistinguishable from the
ceiling band. On MAE the probe is genuinely better than one annotator predicting another: 1.53
versus 2.03, a paired difference of $-0.49$ with CI $[-0.69, -0.29]$. The correct reading is that
the probe lands in the same band as human consensus, not that the machine beats the annotators.

\begin{figure}[t]
\centering
\includegraphics[width=0.98\textwidth]{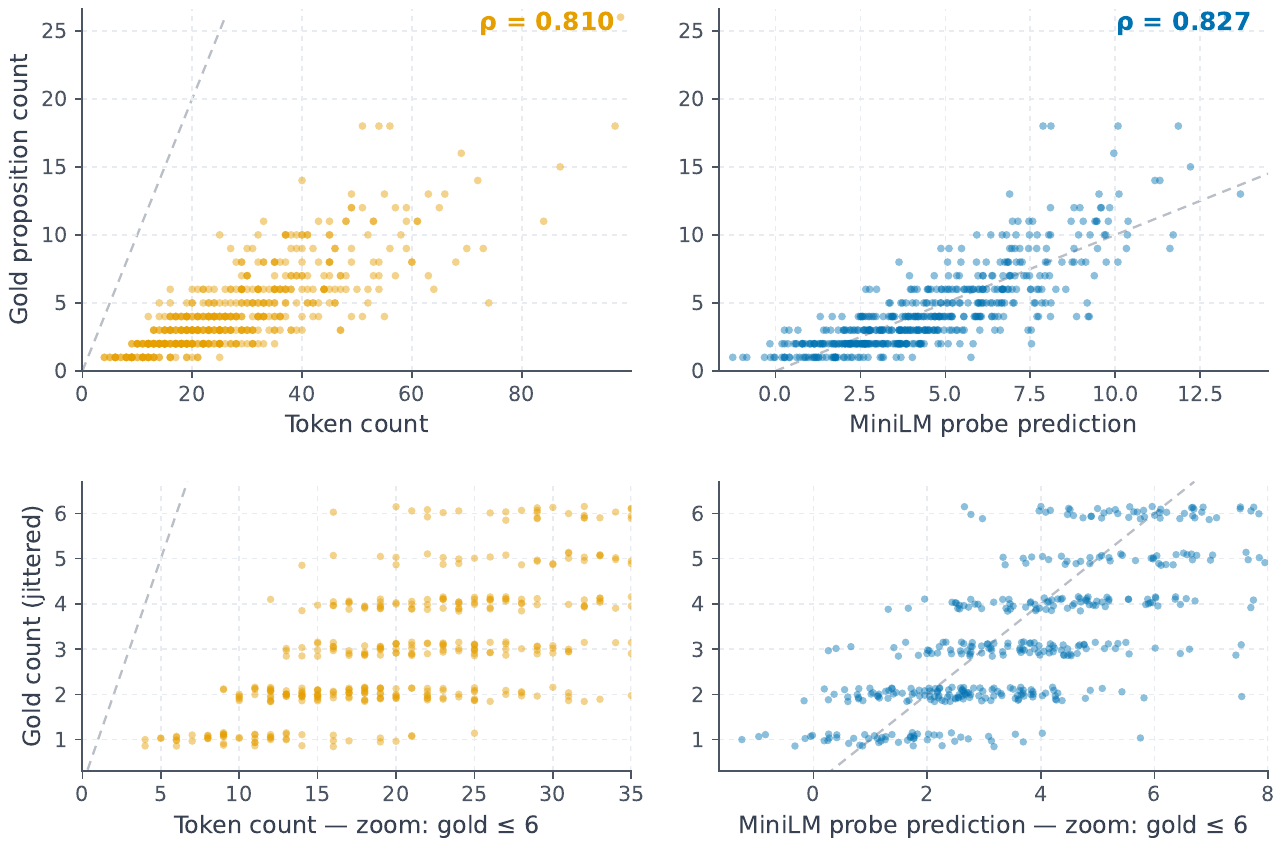}
\caption{The length confound and the probe signal, per test sentence. Left: the token count
already ``counts'' at $\rho=0.810$. Right: the MiniLM probe at $\rho=0.827$, the only predictor
that wins the raw comparison while retaining length-independent signal. Dashed line: $y=x$.}
\label{fig:scatter}
\end{figure}

\subsection{A retraction, kept visible}
\label{sec:retraction}

Before the formal protocol, we ran a pilot on 48 sentences with five decomposition models, using
leave-one-out cross-validation and whitening fitted on the evaluation sentences themselves. It
produced three plausible conclusions: a large encoder-generation gap, with the old encoder at
partial $\rho$ 0.14 against 0.58 for the modern one; whitening as a requirement, lifting 0.44 to
0.58; and the failure of the 2021 encoder. The setup was broken. With $n\ll d$ and
evaluation-fitted whitening, the held-out sentence is nearly orthogonal to the training set in
whitened space, and the minimum-norm leave-one-out ridge solution geometrically memorizes the
label, so the leave-one-out prediction approaches the label itself and cross-validation rewards
the degeneracy. Both closed-form and explicit-refit variants reproduce it, and we retracted all
pilot regression numbers. The formal protocol above, with a large training set, train-only
whitening, a disjoint test split, and inner cross-validation only, reverses all three pilot
conclusions, as the next section shows. We keep the episode in the paper because it is a clean,
replicable example of a probing pitfall that produces plausible numbers rather than obviously
wrong ones.

\subsection{Robustness: encoders, quantization, dimensions}
\label{sec:robust}

\begin{figure}[t]
\centering
\includegraphics[width=0.98\textwidth]{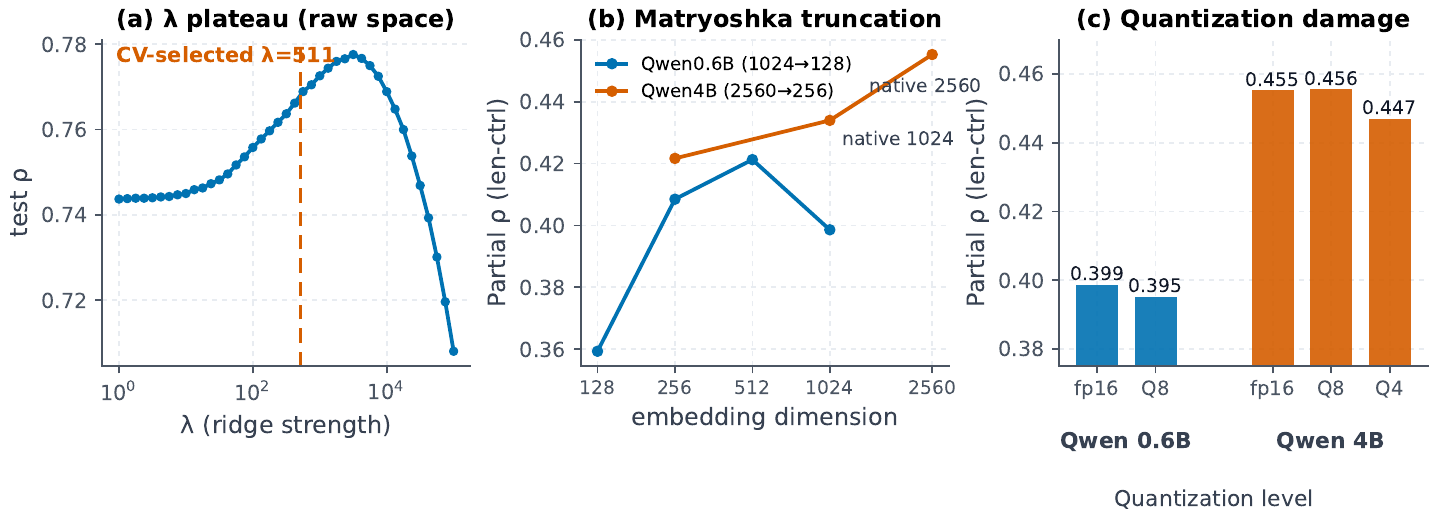}
\caption{Robustness of count decodability. (a) The selected ridge penalty sits on a plateau two
orders of magnitude wide. (b) Matryoshka truncation: the 0.6B encoder at 512 dimensions exceeds
its native dimension, and both families retain pair accuracy of at least 0.68 at 256 dimensions.
(c) Quantization: INT8 is lossless and INT4 costs 0.009 partial $\rho$.}
\label{fig:robust}
\end{figure}

Three hypotheses remained open after the main comparison, each inherited from the pilot or from
the literature: does encoder generation matter, does quantization damage the signal, and how much
dimensionality does the signal need? Table~\ref{tab:robust} and Figure~\ref{fig:robust} answer
all three.

\begin{table}[t]
\centering\small
\begin{tabular}{llcc}
\toprule
Sweep & Configuration & \partialrho & \pairacc \\
\midrule
\multirow{6}{*}{Encoder zoo}
 & MiniLM-L6 384-d & \textbf{0.490} & 0.698 \\
 & Qwen 0.6B fp16 (1,024-d) & 0.399 & 0.669 \\
 & Qwen 0.6B @512-d & 0.421 & 0.691 \\
 & Qwen 0.6B @256-d & 0.409 & 0.686 \\
 & Qwen 0.6B @128-d & 0.359 & 0.672 \\
 & Qwen 4B fp16 (2,560-d) & 0.455 & \textbf{0.710} \\
\midrule
\multirow{2}{*}{Quantization}
 & Qwen 0.6B INT8 & 0.395 & 0.670 \\
 & Qwen 4B INT8 / INT4 & 0.456 / 0.447 & 0.709 / 0.706 \\
\midrule
\multirow{2}{*}{Matryoshka}
 & Qwen 4B @1,024-d & 0.434 & 0.693 \\
 & Qwen 4B @256-d & 0.422 & 0.687 \\
\midrule
\multirow{3}{*}{\shortstack[l]{Whitening\\(train-fitted)}}
 & MiniLM whitened & 0.495 & 0.697 \\
 & Qwen 0.6B whitened & 0.358 & 0.672 \\
 & Qwen 4B whitened & 0.348 & 0.663 \\
\bottomrule
\end{tabular}
\caption{Robustness sweep on the test split (raw-space numbers except the whitening block).
INT8 (GGUF \texttt{q8\_0}) is lossless: the 4B raw $\rho$ is 0.7895 under INT8 against 0.7891
under fp16. INT4 (\texttt{q4\_K\_M}) costs 0.009 partial $\rho$. The 0.6B encoder at 512
dimensions exceeds its native 1,024. Whitening is unnecessary and increasingly harmful for modern
encoders.}
\label{tab:robust}
\end{table}

\paragraph{The generation hypothesis is false.}
The 2021 MiniLM, with 22M parameters and mean pooling, outperforms the 2025 Qwen3 4B on raw
$\rho$ at 0.827 versus 0.789, on partial $\rho$ at 0.490 versus 0.455, and on MAE at 1.36 versus
1.48, with heavily overlapping CIs; the 4B leads only on pair accuracy at 0.710 against 0.698,
again with overlapping CIs. Count decodability is a broad property of sentence embeddings rather
than a modern-encoder capability. Capacity does help within the Qwen family, from 0.6B to 4B on
partial $\rho$ 0.399 to 0.455 and pair accuracy 0.669 to 0.710, so the honest statement is that
generation is not the determining variable while within-family scale contributes modestly.

\paragraph{Quantization damage is negligible.}
INT8 is statistically indistinguishable from fp16 in every metric, and INT4 costs 0.003 raw
$\rho$ and 0.009 partial $\rho$. This is consistent with the 4-bit inference scaling results of
\citet{dettmers2023int4}, and it means that a 2.5\,GB INT4 model screens granularity as well as
the 8\,GB fp16 original, so the application is deployable almost anywhere.

\paragraph{Dimensions compress gracefully.}
Through the encoder's native Matryoshka projection \citep{kusupati2022matryoshka}, the 0.6B
encoder at 512 dimensions exceeds its native 1,024, at partial $\rho$ 0.421 against 0.399 and
pair accuracy 0.691 against 0.669, and both families retain pair accuracy of at least 0.68 at 256
dimensions.

\paragraph{Whitening: from requirement to liability.}
Train-fitted whitening is neutral for MiniLM at $+0.005$ partial $\rho$, mildly harmful for 0.6B
at $-0.04$, and clearly harmful for 4B at $-0.11$. The pilot-era whitening requirement was an
artifact of evaluation-set fitting; on the formal protocol the raw space is the right default,
and the damage grows as encoders become more contrastively trained.

\section{Analysis: Where the Probe Fails}
\label{sec:analysis}

The probe is not uniformly accurate, and the way it fails is as informative as the way it
succeeds. This section dissects the residual error by sentence type and then closes the loop with
an application to anchor repair.

\begin{table}[t]
\centering\small
\begin{tabular}{lcccccc}
\toprule
Stratum & $n$ & Mean gold count & MAE & Bias & $\rho$ (probe) & $\rho$ (tokens) \\
\midrule
quantifier & 328 & 5.05 & 1.44 & $-$0.004 & 0.811 & 0.776 \\
negation & 22 & 4.36 & 1.06 & $+$0.445 & 0.904 & 0.901 \\
modal & 22 & 3.77 & 0.97 & $-$0.044 & 0.874 & 0.806 \\
\textbf{long} ($\geq$36 tok) & 136 & \textbf{8.14} & \textbf{1.92} & $\mathbf{-0.686}$ & 0.724 & \textbf{0.456} \\
plain (control) & 162 & 2.91 & 1.22 & $+$0.232 & 0.747 & 0.741 \\
\bottomrule
\end{tabular}
\caption{Stratified analysis of the MiniLM probe. Strata are heuristically labeled and
non-exclusive; \emph{plain} contains sentences with none of the markers. Error concentrates in
long sentences as systematic under-splitting, with bias $-0.69$ for MiniLM, $-0.74$ for 0.6B, and
$-0.97$ for 4B. The token count's within-stratum $\rho$ collapses to 0.456 on the long stratum.}
\label{tab:strata}
\end{table}

\begin{figure}[t]
\centering
\includegraphics[width=0.98\textwidth]{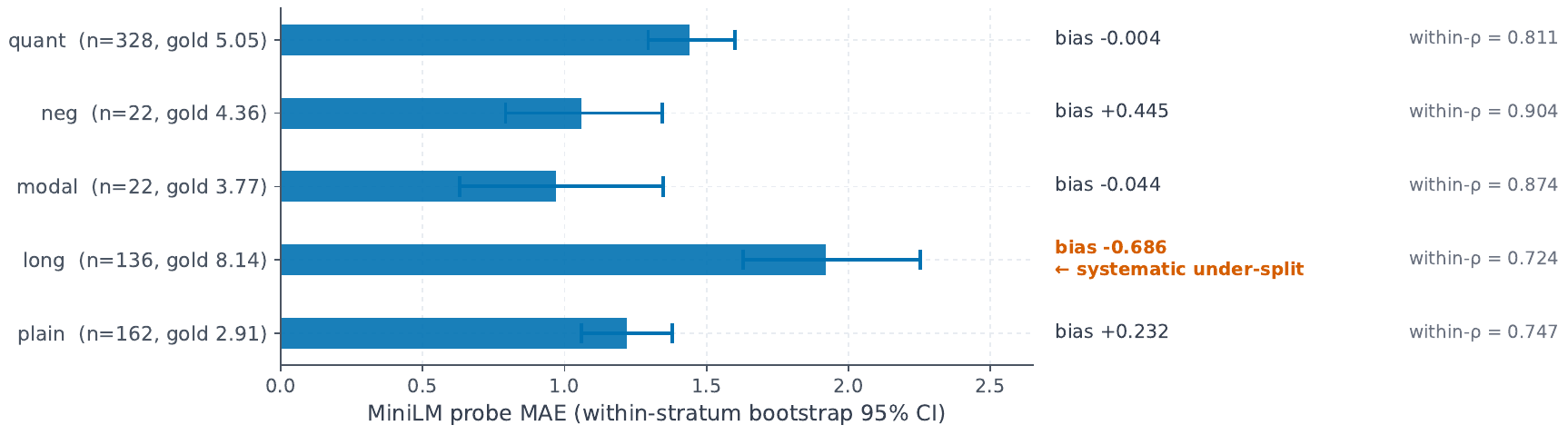}
\caption{Stratified MAE with bootstrap CIs and signed bias. The prediction error is structured:
long sentences are systematically under-split, while quantifier-bearing sentences are perfectly
calibrated on average.}
\label{fig:strata}
\end{figure}

\begin{figure}[t]
\centering
\includegraphics[width=0.98\textwidth]{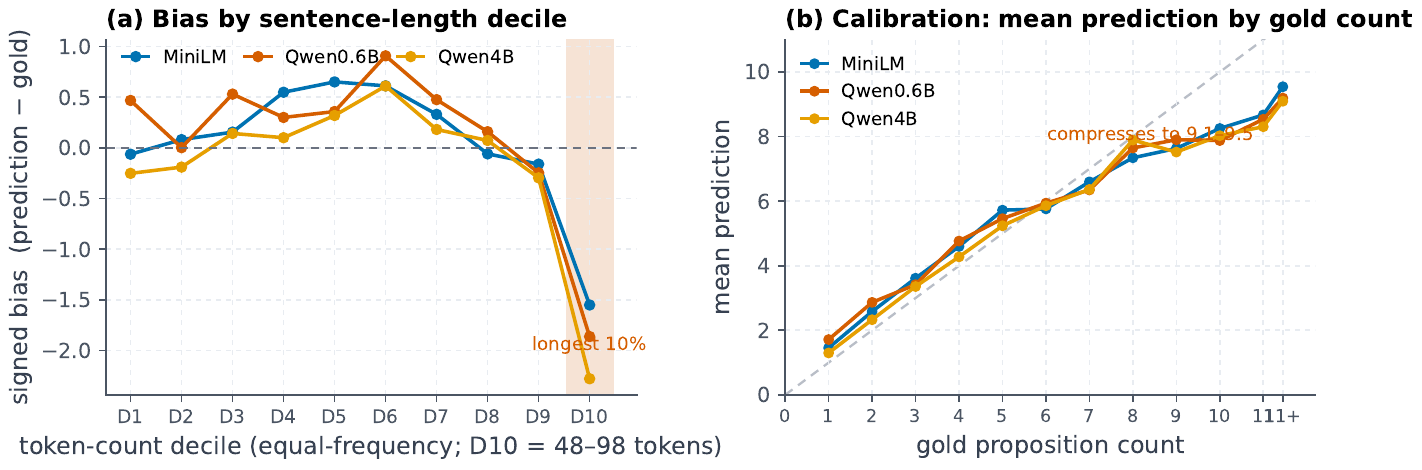}
\caption{Error structure of the three probes on the test split. \emph{Left}: signed bias (mean
prediction minus gold count) by sentence-length decile; the shaded band marks the longest 10\%
of sentences (48--98 tokens), where under-splitting concentrates and reaches $-1.5$ to $-2.3$
across probes. \emph{Right}: calibration, the mean prediction by gold count against the identity
diagonal; mean predictions track the diagonal through gold counts of about five and then fall
progressively behind, compressing to 9.1--9.5 for gold counts of eleven or more.}
\label{fig:errorstruct}
\end{figure}

Table~\ref{tab:strata} and Figure~\ref{fig:strata} stratify test sentences by heuristic markers:
quantifiers or numerals, negation cues, modals, length at or above the 75th percentile of 36
tokens, plus a \emph{plain} control stratum with none of these markers. Three observations
follow.

First, the residual error is a long-sentence under-splitting bias. On the long stratum, whose
mean gold count is 8.14, all three probes under-predict, with bias $-0.69$ for MiniLM, $-0.74$
for 0.6B, and $-0.97$ for 4B, and MAE rising to 1.9--2.1. Figure~\ref{fig:errorstruct} sharpens
the localization. Disaggregated by length decile (left panel), biases hover between $-0.26$ and
$+0.97$ over the first nine deciles, mildly over-predicting at mid lengths, and collapse only in
the longest decile (48--98 tokens), where they reach $-1.53$, $-1.89$, and $-2.27$ respectively.
The calibration view (right panel) suggests the mechanism: mean predictions track the identity
diagonal through gold counts of about five and then fall progressively behind, so that sentences
with gold counts of eleven or more receive mean predictions of only 9.1--9.5. The failure is
specific and directional rather than noise: the probes' count estimates compress as sentences
grow.

Second, decision-type markers are not the problem. The negation and modal strata are the most
accurate, with MAE around 1.0 and within-stratum $\rho$ of 0.87--0.90, although their sizes of 22
sentences each warrant caution. Quantifier and numeral sentences, 61\% of the test set, are
perfectly calibrated on average at bias $-0.004$.

Third, the length signal also collapses exactly where it matters. Within the long stratum, the
token count's $\rho$ falls to 0.456, and on short plain sentences every predictor weakens
because count variance is low. The regime where an external count estimate would be most
valuable, namely long and information-dense sentences, is precisely where all cheap predictors
are weakest, which bounds the application story of Section~\ref{sec:implications}.

\paragraph{Probe as anchor repair.}
In a complementary validity experiment, we scored the human gold segmentation itself with a
proposition-quality index whose granularity channel compares produced counts against the symbolic
anchor g\_dav. Replacing g\_dav with the probe's prediction moves granularity agreement from
0.620 to 0.717 and the red-flag rate from 48\% to 27\%, while a perfect anchor, namely the human
count, yields 1.0 and 0\%. The probe thus works as a second opinion that repairs symbolic
anchors, not as a replacement for human judgment.

\section{Implications}
\label{sec:implications}

Three practical uses follow from the two branches, in each case because one direction is solved
and the other is not.

Verification is accurate but pairwise, while prediction is coarse but constant-time per sentence,
so a quality-control pipeline should use the probe to budget and route: sentences whose predicted
count implies high decomposition cost, or whose predicted-to-produced ratio drifts, are routed to
the NLI judge. The probe's cost profile, INT4-quantized at 256 dimensions, makes it essentially
free at inference.

Downstream consumers of decomposition, from retrieval-unit budgeting to per-split pricing, care
about the expected count before running the decomposer. An MAE of 1.4--1.9 with a known
long-sentence bias is often enough for such budgeting decisions, and the direction of the bias,
under-splitting, is now known.

Finally, the pair of findings revises what we assume embeddings contain. The sentence embedding
does not linearly contain its propositions' identities, yet it does linearly contain a decent
estimate of how many propositions there are. Granularity, unlike faithfulness, is a broadly
distributed and robustly decodable property, present already in 2021-era encoders and surviving
aggressive compression.

\section{Limitations}
\label{sec:limitations}

Seven limitations bound the claims. First, linear probes are lower bounds: nonlinear readouts
could exceed a partial $\rho$ of 0.49, and our claim is about what is linearly available. Second,
all quantitative conclusions come from one dataset and one domain, English Wikipedia; transfer to
ESG disclosure text is planned but not yet run. Third, proposition count is annotator-defined:
our ceiling is agreement, not truth, and PropSegmEnt's span-style propositions differ from
self-contained-proposition schemes, so cross-scheme generalization is untested. Fourth, the NLI
judge is not neutral: residual false positives concentrate in context-true propositions, and a
document-aware membership definition would shift the operating point. Fifth, the negation and
modal strata contain only 22 sentences each. Sixth, our CPIDR and g\_dav are faithful
reimplementations rather than the original toolchains, so conclusions are about the
symbolic-anchor family rather than any specific implementation. Seventh, all probing uses frozen
sentence encoders; LLM hidden states, which a 4B embedding model only partially proxies, may
contain more.

\section{Ethics and License}
\label{sec:ethics}

All experiments use publicly available resources. PropSegmEnt \citep{chen2023propsegment}
redistributes sentence texts from English Wikipedia under CC BY-SA 3.0; we release only derived
statistics such as counts, predictions, and scores, and at most short illustrative sentences with
attribution. The dataset repository does not ship a license file, so we pin the exact commit
(\texttt{d5172f956e331540524dae21f0a41c9de906ac08}) for reproducibility. The NLI judge,
mDeBERTa-v3 fine-tuned on multilingual-NLI-2mil7, is MIT-licensed; all-MiniLM-L6-v2 and
Qwen3-Embedding, in fp16 and GGUF variants, are Apache-2.0-licensed. No personal or sensitive
data is used.

Two interpretive caveats apply. First, proposition count is a human construct: our ceiling is
inter-annotator agreement, $\rho=0.735$ with pooled annotator-pair MAE 2.03 on the test split,
not an objective truth, and the span-based annotation style differs from self-contained
proposition schemes, which is why we triangulate across anchors and report it as a limitation.
Second, the membership judge inherits the biases of its NLI training data; our false-positive
analysis, which surfaces context-true definitional propositions judged entailed via coreference,
characterizes rather than eliminates this bias. The count probe is a measurement tool for
granularity screening, and we do not recommend it as a stand-alone decision maker for content
moderation or deletion.

\section{Conclusion}
\label{sec:conclusion}

Proposition decomposition admits two quality-control questions, and they are not equally open.
Whether a proposition belongs to its sentence is verifiable today with a discriminative NLI judge
at 93\% accuracy, while vector geometry cannot verify at all. How many propositions a sentence
will yield is only partially predictable: a linear ridge probe on sentence embeddings carries
genuine length-independent signal, matches the inter-annotator agreement ceiling, and survives
quantization, dimension truncation, and encoder generation, yet systematically under-splits long
sentences. The two halves of the relation between a sentence and its propositions thus live in
different representational spaces, faithfulness in symbolic inferential space and granularity
partially in embedding space, and each half now has a quantified, bounded answer. Future work
includes domain transfer to ESG disclosure text, nonlinear and LLM-hidden-state probes,
document-aware membership, and threshold-free consensus anchoring across decomposition models.

\paragraph{Reproducibility.}
All experiments run on a single Apple Silicon workstation. Ollama provides fp16 embeddings with
SHA-256 disk caches; MiniLM and the NLI judge run on the Metal GPU; the full encoder zoo
including the 4B quantized variants completes in 43 minutes. The formal protocol is
deterministic and was verified to reproduce its headline numbers bit-exactly, and build-time
cross-checks on dataset sizes, histogram conservation, and recomputed correlations accompany
every generated figure. Statistics are computed with SciPy \citep{virtanen2020scipy}.

\bibliographystyle{plainnat}
\bibliography{refs}

\appendix

\section{Implementation Details}
\label{app:impl}
Embeddings are unit-normalized and cached by the SHA-256 hash of the sentence text; whitening,
when used, is fitted on train only with eigenvalues clipped at $10^{-8}$. Ridge features are
standardized; the $\lambda$ grid spans $10^{[0,5]}$ in 25 points with 5-fold cross-validation
inside train and a deterministic fold assignment with seed 0. Bootstrap uses 2{,}000 resamples
for the main-table CIs and 1{,}000 for the zoo, sentence-level for $\rho$, MAE, and partial
$\rho$, and pair-level for pair accuracy. Partial correlation is the Pearson correlation of rank
residuals after regressing each variable on token count. Membership pairs are sampled
deterministically with seed 42, with two negatives per sentence per type where available. Strata
are defined as follows: the quantifier stratum contains closed-class quantifiers, quantity
phrases such as \emph{at least} or \emph{more than}, or any digit; the negation stratum contains
\emph{not}, \emph{never}, \emph{no}, \emph{none}, \emph{neither}, \emph{nor}, \emph{without}, or
a contracted \emph{n't}; the modal stratum contains \emph{will}, \emph{would}, \emph{can},
\emph{could}, \emph{might}, \emph{must}, \emph{should}, \emph{shall}, or a lowercase \emph{may},
which avoids the month; the long stratum is a token count at or above the test-split 75th
percentile of 36.

\section{Retraction Record}
\label{app:retraction}
The retracted pilot numbers, from 48 sentences with five decomposition models using
leave-one-out ridge and whitening fitted on evaluation sentences, are: MiniLM $\rho=0.46$ and
partial $\rho$ 0.14; Qwen3-0.6B $\rho=0.87$ and partial $\rho$ 0.58; and a whitening improvement
from 0.44 to 0.58. The mechanism is that with $n\ll d$ and self-fitted whitening, the
leave-one-out point is near-orthogonal to the training set, and the minimum-norm ridge solution
interpolates the label at the held-out point, which cross-validation then selects for. The
degeneracy reproduces under both the closed-form leverage formula, after correcting a
shape-alignment bug that crashes for $n>d$ and silently mis-scores for $n<d$, and explicit
refitting. No pilot regression number is cited anywhere in this paper's results; geometry pilots
on span residual, cosine saturation, and spectral dimension were recomputed under the formal
protocol or are labeled as pilot-scale where cited.

\section{Full Encoder-Zoo Results}
\label{app:zoo}
The complete sweep is stored in \texttt{out/p1c\_e3e4\_zoo.json}: 11 configurations with
bootstrap CIs, including whitened spaces and all Matryoshka points, of which
Table~\ref{tab:robust} shows the raw-space highlights. Embedding wall-times per configuration
are approximately 14 seconds for MiniLM, 107 seconds for any Qwen 0.6B variant regardless of
precision or dimension, and 400--460 seconds for Qwen 4B.

\end{document}